\documentclass[a4paper,twoside]{article}

\usepackage{epsfig}
\usepackage{subcaption}
\usepackage{calc}
\usepackage{amssymb}
\usepackage{amstext}
\usepackage{amsmath}
\usepackage{amsthm}
\usepackage{multicol}
\usepackage{pslatex}
\usepackage{apalike}
\usepackage{soul}
\usepackage{algorithm2e}
\usepackage{graphicx}
\graphicspath{{figures/}}
\usepackage[bottom]{footmisc}
\usepackage{SCITEPRESS}     

\begin{document}

\title{DevPrompt: Deviation-Based Prompt Learning for One-Normal Shot Image Anomaly Detection}

    
\author{\authorname{Morteza Poudineh\sup{1,2} and Marc Lalonde\sup{2}\orcidAuthor{0000-0002-1183-6378}}
\affiliation{\sup{1}Department of Electrical and Computer Engineering, Concordia University,  Montreal, QC H3G 1M8, Canada }
\affiliation{\sup{2}R\&D Dept., Computer Research Institute of Montreal (CRIM),   405 Ogilvy Ave., \#101, Montreal, Qc, Canada}
\email{morteza.poudineh@concordia.ca, Marc.lalonde@crim.ca}
}



\abstract{Few-normal shot anomaly detection (FNSAD) aims to detect abnormal regions in images using only a few normal training samples, making the task highly challenging due to limited supervision and the diversity of potential defects. Recent approaches leverage vision-language models such as CLIP with prompt-based learning to align image and text features. However, existing methods often exhibit weak discriminability between normal and abnormal prompts and lack principled scoring mechanisms for patch-level anomalies. We propose a deviation-guided prompt learning framework that integrates the semantic power of vision-language models with the statistical reliability of deviation-based scoring. Specifically, we replace fixed prompt prefixes with learnable context vectors shared across normal and abnormal prompts, while anomaly-specific suffix tokens enable class-aware alignment. To enhance separability, we introduce a deviation loss with Top-K Multiple Instance Learning (MIL), modeling patch-level features as Gaussian deviations from the normal distribution. This allows the network to assign higher anomaly scores to patches with statistically significant deviations, improving localization and interpretability. Experiments on the MVTecAD and VISA benchmarks demonstrate superior pixel-level detection performance compared to PromptAD and other baselines. Ablation studies further validate the effectiveness of learnable prompts, deviation-based scoring, and the Top-K MIL strategy.}

\onecolumn  \maketitle \normalsize \setcounter{footnote}{0} \vfill

\section{Introduction}

Few-normal shot anomaly detection (FNSAD) considers scenarios with only a handful of normal samples for training, reflecting realistic industrial constraints \cite{patchcore,winclip,promptad,visap1,visap2,visap3}. Pioneering works such as PatchCore, WinCLIP, and PromptAD emphasize patch-level features, prompt engineering, and language-guided representations. Nonetheless, challenges remain: heterogeneity of anomalies requires generalization to unseen cases, and the semantic gap between normality and abnormality needs explicit modeling for fine-grained detection.

To address these limitations, deviation-based learning \cite{devnet} has been explored, where anomaly scores are derived from statistical deviations of patch features relative to a Gaussian prior. This provides interpretability and discriminability. Combining deviation-based scoring with prompt learning allows capturing semantic context while enforcing statistical separation at the patch level.

State-of-the-art FNSAD methods can be categorized into (i) prompt-driven representation learning, focusing on effective textual templates for discriminative embedding spaces, and (ii) deviation-guided anomaly scoring, which enforces statistical constraints to distinguish anomalies. The former enhances semantic generalization, while the latter improves interpretability and robustness.

Building on these insights, we propose a framework integrating learnable prompts with deviation loss \cite{devnet} for patch-level scoring. Learnable prompts adaptively represent both normal and abnormal contexts, while Top-K Multiple Instance Learning (MIL) aggregates patch-level deviations, improving localization. This unified approach combines semantic alignment with statistical deviation, advancing generalizable and interpretable few-normal shot anomaly detection. 

Our contributions are as follows:
\begin{itemize}
\item A unified framework integrating prompt-based anomaly detection with deviation-guided scoring.
\item Learnable prompts representing normal and abnormal contexts while retaining class-specific cues.
\item Patch-level deviation loss for improved discriminability and interpretability.
\item Top-K MIL aggregation to handle sparse anomalies and enhance localization.
\item Demonstrated improved generalization to unseen anomaly types with minimal normal samples.
\end{itemize}

The rest of the paper is organized as follows. Section \ref{Related work} reviews anomaly detection literature. Section \ref{sec:methodology} details our prompt-based anomaly detection framework with MIL-driven deviation loss. Section \ref{experiment} presents experimental results. Section \ref{conclusion} concludes and discusses future directions.
\section{Related works}
\label{Related work}
Anomaly detection has been explored in unsupervised, supervised, and few-shot settings, or limited labeled anomalies. In following we review some of these works.
\subsection{Unsupervised and Few-shot Anomaly Detection}
Most anomaly detection methods operate under unsupervised or weakly supervised settings using only normal data for training. Early approaches rely on reconstruction-based models such as autoencoders \cite{autoencoder1,autoencoder2} and generative methods including AnoGAN \cite{anogan} and GANomaly \cite{ganomaly}, where anomalies are detected via reconstruction errors or latent deviations. Self-supervised learning \cite{geometric-transform,jigsaw} and one-class classification methods such as Deep SVDD \cite{deep_svdd} further aim to learn compact normal representations. Extensions incorporating memory mechanisms \cite{memad}, knowledge distillation \cite{kdad}, or limited anomaly supervision \cite{deep_sad}, metric learning \cite{deep_metric1,deep_metric2}, and MIL-based frameworks \cite{mil_video1,mil_video2} improve discriminability in few-shot settings. DevNet \cite{devnet} introduces a deviation-based objective that directly optimizes anomaly separation without relying on reconstruction quality. Despite these advances, many methods struggle with subtle or spatially localized anomalies and often lack interpretability in industrial scenarios.
\subsection{CLIP-based Anomaly Detection}
Recent works leverage vision--language models such as CLIP \cite{clip} for anomaly detection via prompt-based similarity modeling. CLIP-AD \cite{clip_ad} enables zero-shot detection by aligning images with textual descriptions of normal and abnormal states, while PromptAD \cite{promptad} further learns optimal prompts from few-shot normal samples. Other approaches explore prompt tuning and in-context learning strategies \cite{tall,inctrl} to improve generalization. However, existing CLIP-based methods primarily rely on raw similarity scores, which limits their ability to explicitly model statistical deviations between normal and anomalous regions, especially for fine-grained or localized defects.
\section{Methodology}
\begin{figure*}
    \centering
    \includegraphics[width=.8\linewidth]{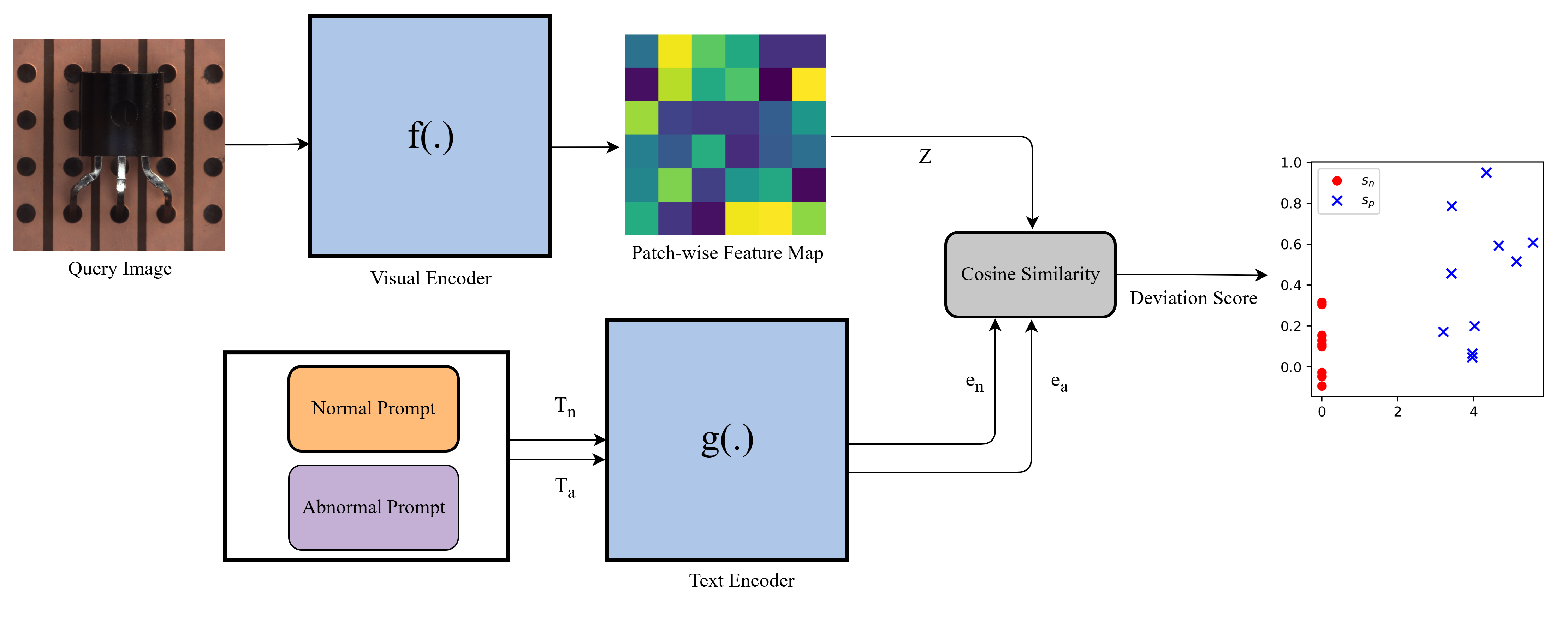}
    \caption{Visualization of patch-wise anomaly scores computed using the proposed deviation-guided prompt learning approach. Higher deviation scores correspond to more anomalous regions.}
    \label{fig:placeholder}
\end{figure*}
\label{sec:methodology}
In this section, we present our proposed framework for few-normal shot anomaly detection.
The method is designed to capture discriminative alignment between image patches and textual prompts while providing a statistically grounded measure of deviation for anomaly scoring.
The overall pipeline consists of four key stages: (i) visual--textual embedding, (ii) prompt construction with learnable context, (iii) deviation-based anomaly scoring, and (iv) optimization with deviation-aware loss.

\subsection{Problem Definition}

Let $\mathcal{X} = \{x_i\}_{i=1}^N$ denote the input images, where each $x_i$ may contain normal or anomalous regions.
Each image is decomposed into $P$ non-overlapping patches, denoted as $x = \{x_p\}_{p=1}^P$.
A dual-encoder model is employed, where the visual encoder $f(\cdot)$ extracts patch-level embeddings and the textual encoder $g(\cdot)$ projects tokenized prompts into the same embedding space:
\begin{equation}
\begin{aligned}
    z_p &= f(x_p) \in \mathbb{R}^d, \\[5pt]
    e_a &= g(T_a) \in \mathbb{R}^d, \qquad e_n &= g(T_n) \in \mathbb{R}^d
    \vspace{210pt}
\end{aligned}
\end{equation}
where $z_p$ is the embedding of patch $p$, $e_a$ and $e_n$ are the representations of textual prompt $T_a$ and $T_n$ respectively. 
\subsection{Prompt Construction with Learnable Context}
Traditional prompt-based approaches, including PromptAD \cite{promptad}, rely on fixed textual tokens, such as manually designed prefixes and suffixes, to represent normal and abnormal contexts. While this design provides stability and class-specific guidance, it can limit discriminability when anomaly patterns are subtle or diverse. Our work builds on this framework by retaining the fixed prompt structure but introducing deviation-based scoring at the patch level, which explicitly enforces statistical separation between normal and abnormal features. This allows the model to leverage the stable prompt representation while improving anomaly localization and robustness in few-shot scenarios.

The normal and abnormal prompts are constructed as
\begin{equation}
    T_n = [C; \text{suffix}_n], 
    \qquad
    T_a = [C; \text{suffix}_a],
\end{equation}

where $\text{suffix}_n$ and $\text{suffix}_a$ are fixed, semantically meaningful tokens corresponding to class-specific and anomaly-specific textual descriptions. For instance, for a given object class $c =$ ``bottle'', the normal and abnormal prompts may be expressed as  
``a photo of a normal bottle'' and ``a photo of a defective bottle'', respectively.
This design preserves interpretability while enabling the model to adaptively learn discriminative context.
\subsection{Patch-to-Prompt Alignment}
Given patch embeddings $z_p$ and prompt embeddings $e_n, e_a$, we compute cosine similarities:
\begin{equation}
    s_n(p) = \frac{z_p \cdot e_n}{\|z_p\| \, \|e_n\|}, 
    \qquad
    s_a(p) = \frac{z_p \cdot e_a}{\|z_p\| \, \|e_a\|}.
\end{equation}
Here, $s_n(p)$ measures the alignment of patch $p$ with normal semantics, while $s_a(p)$ captures its alignment with abnormal cues.
Collecting these scores over all patches provides a fine-grained representation of how an image relates to both normality and abnormality.
\begin{algorithm}[t]
\caption{Proposed Few-normal shot Anomaly Detection Framework}
\label{alg:fsad}
\KwIn{Training images $\mathcal{X}_{train}$ (normal samples)}
\KwOut{Image-level anomaly score $D(x)$}

\textbf{Training Phase:}\;
\For{each normal image $x_i \in \mathcal{X}_{train}$}{
    Divide $x_i$ into $P$ patches: $\{x_{i,p}\}_{p=1}^P$\;
    Extract patch embeddings: $z_{i,p} = f(x_{i,p})$\;
}
Initialize learnable context vectors $C = [c_1, ..., c_k]$\;
Construct normal and abnormal prompts:
$T_n = [C; \text{suffix}_n], \quad T_a = [C; \text{suffix}_a]$\;
Compute prompt embeddings: $e_n = g(T_n), \quad e_a = g(T_a)$\;
\For{each patch embedding $z_{i,p}$}{
    Compute cosine similarities:
    $s_n(p) = \frac{z_{i,p} \cdot e_n}{\|z_{i,p}\|\|e_n\|}, \quad
     s_a(p) = \frac{z_{i,p} \cdot e_a}{\|z_{i,p}\|\|e_a\|}$\;
}
Estimate Gaussian prior $(\mu, \sigma)$ from $s_n(p)$\;
Compute deviation scores: $d(p) = \frac{|s_a(p)-\mu|}{\sigma}$\;
Optimize the model by minimizing:
\[
\mathcal{L} =  \lambda \mathcal{L}_{dev}
\]

\textbf{Return:} trained parameters $\theta$ ;
\end{algorithm}
\subsection{MIL-Driven Deviation Loss}
Cosine similarities alone are insufficient to reliably distinguish anomalies, because the distributions of similarity scores for normal and abnormal patches often overlap. To address this, we introduce a deviation-based scoring mechanism grounded in statistical modeling. 

We assume that the distribution of normal patch-to-normal prompt similarities follows a Gaussian distribution:
\begin{equation}
    s_n(p) \sim \mathcal{N}(\mu, \sigma^2),
\end{equation}
where $\mu$ and $\sigma$ are estimated \hl{ following DevNet} \cite{devnet}\hl{, where a fixed standard Gaussian prior is defined, and a large set of reference values is sampled from this distribution to determine the $\mu$ and $\sigma$.}

For each patch, the deviation score is defined as:
\begin{equation}
    d(p) = \frac{|s(p) - \mu|}{\sigma}.
\end{equation}
Intuitively, $d(p)$ quantifies how strongly a patch deviates from the expected normal similarity, with larger values indicating potential anomalies. To robustly aggregate patch-level evidence, we adopt a Top-$K$ \textit{Multiple Instance Learning (MIL)} strategy inspired by \cite{devnet}.
We define the set of top-$K$ patches with the largest deviations as
\begin{equation}
    \mathcal{M}(x) = \text{Top-K}\{ d(p) \mid p = 1, \dots, P\}.
\end{equation}
This selection is motivated by two observations: (i) anomalous images contain normal background patches, and (ii) normal images may contain ``hard'' patches resembling anomalies. Learning from only the top-$K$ patches ensures that the model focuses on the most informative regions for both anomaly detection and hard negative mining. Finally the aggregated image-level deviation is computed as
\begin{equation}
    D(x) = \frac{1}{K} \sum_{m \in \mathcal{M}(x)} m.
\end{equation}

We then define the \textit{deviation loss} to encourage a statistically significant separation between normal and anomalous images:
\begin{equation}
\mathcal{L}_{dev}(x_i, y_i) = \lambda *
(1 - y_i) \, |D(x_i)| + 
y_i \, \max\big(0, a - D(x_i) \big),
\end{equation}
where $y_i = 1$ if $x_i$ is anomalous, $0$ otherwise, and $a$ is a confidence parameter (Z-score threshold) controlling the margin between normal and anomalous deviations. Following \cite{devnet}, we set $a = 5$ for high statistical significance. The coefficient $\lambda$ represents the deviation scaling factor, we empirically set it to $1$ in our experiments. This formulation ensures that: (i) The deviation of normal images are pushed to have deviations close to zero. (ii) Anomalous images are forced to exceed a threshold $a$. 
\section{Experimental Results}
\label{experiment}
\begin{table}[h!]
\centering
\caption{MVTecAD Dataset: Class-wise AUROC (\%). Bold indicates improvement over the baseline PromptAD.}
\label{tab:mvtec_results}
\begin{tabular}{lccc}
\hline
Class & WinCLIP&PromptAD & Proposed \\
\hline
Bottle & 96.8& 97.31 & 97.28 \\
Cable & 93.6& 94.10 & 93.79 \\
\textbf{Capsule} &96.04 &95.60 & \textbf{96.25} \\
Carpet & 99.1&99.54 & 99.54 \\
Grid & 96.3&97.99 & 97.97 \\
\textbf{Hazelnut} & 97.76&98.21 & \textbf{98.27} \\
Leather & 99.2& 99.34 & \textbf{99.36} \\
\textbf{Metal Nut} &89.5 &91.65 & \textbf{92.67} \\
\textbf{Pill} & 93.8& 93.63 & \textbf{95.26} \\
Screw & 95.9&95.19 & 94.11 \\
\textbf{Tile} & 95.7&96.05 & \textbf{96.06} \\
\textbf{Toothbrush} & 97&98.63 & \textbf{98.67} \\
\textbf{Transistor} & 84.3&85.65 & \textbf{85.66} \\
Wood & 94.8&95.44 & 95.29 \\
\textbf{Zipper} & 91.4&90.77 & \textbf{92.04} \\
\hline
\end{tabular}
\vspace{5pt}
\end{table}
In this section, we present a comprehensive evaluation of the proposed PromptAD framework augmented with deviation-based patch-wise scoring for anomaly detection. All experiments are conducted on the \textbf{MVTecAD} \cite{mvtec_data} and \textbf{VISA} \cite{visa} datasets, with a particular focus on \textbf{pixel-level anomaly score}. The evaluation highlights the effectiveness of patch-level Top-K MIL scoring in capturing fine-grained anomalies. We report class-wise AUROC (\%) and perform sensitivity analayzes to measure the impact of the deviation coefficient $\lambda$, the Top-K parameter $K$ and the confidence parameter $a$ on model performance.
\subsection{Datasets}
\label{sec:datasets}
The \textbf{MVTecAD} dataset consists of 15 industrial object categories (e.g., \textit{bottle, cable, hazelnut, transistor}) with high-resolution images of size approximately $700 \times 900$ pixels. The dataset includes both normal and defective samples, with pixel-level ground truth annotations for anomalies. Training is conducted on \emph{normal} images only, while the test set includes both normal and anomalous images.  

The \textbf{VISA} dataset contains 12 object categories (e.g., \textit{cashew, chewinggum, PCB}), with an average image resolution of approximately $1500 \times 1000$ pixels. Similar to MVTec, VISA provides normal samples for training and both normal and anomalous samples with fine-grained pixel-level annotations for testing.  
\begin{table}[h!]
\centering
\caption{VISA Dataset: Class-wise AUROC (\%) for pixel-level segmentation. Bold indicates improvement over PromptAD.}
\label{tab:visa_results}
\begin{tabular}{lccc}
\hline
Class & WinCLIP&PromptAD & Proposed \\
\hline
\textbf{Candle} & 94.53&94.65 & \textbf{94.69} \\
Capsules & 95.9&94.01 & 94.01 \\
\textbf{Cashew} & 97.9&98.85 & \textbf{98.88} \\
\textbf{Chewinggum} & 99&99.24 & \textbf{99.25} \\
\textbf{Fryum} & 95.33&95.58 & \textbf{95.63} \\
Macaroni1 & 96.8&97.81 & 97.73 \\
\textbf{Macaroni2} &94.87 & 95.39 & \textbf{95.49} \\
\textbf{PCB1} &95.31 &95.72 & \textbf{95.83} \\
PCB2 & 93.04&94.30 & 93.34 \\
PCB3 & 94.34&94.66 & 94.65 \\
PCB4 & 95.3 &94.62 & 94.49 \\
\textbf{Pipe\_Fryum} & 98.92&99.16 & \textbf{99.19} \\
\hline
\vspace{-12pt}
\end{tabular}
\end{table}
For both datasets, we follow the standard train/test split defined in the benchmark \cite{promptad}. Since the focus of this work is on \emph{few-normal-shot anomaly detection}, all experiments are conducted under a \textbf{1-shot setting}, meaning that only one normal image per class is used during training. \hl{The training set contains only $1$ defect-free image per class, and no anomalous samples are used during training, and generalization is achieved by learning patch-level normality statistics.} This setting aligns with the original PromptAD paper and reflects realistic industrial scenarios where abundant normal data may not be available at design time. 

All images are resized to $256 \times 256$ prior to training and evaluation to ensure consistent input dimensions across categories and datasets.

\subsection{Training Setting}
\label{sec:training}
We adopt the exact hyperparameter configuration used in the baseline \cite{promptad} implementation for fair comparison. The backbone network uses pretrained CLIP weights \cite{clip}. No additional finetuning of the backbone is performed; only learnable prompt tokens and deviation-based scoring parameters are optimized. 

Importantly, in contrast to image-level classification, our deviation-based patch scoring and Multiple Instance Learning (MIL) are applied \emph{only at the pixel level}. The rationale behind this design is that anomaly pixel level focuses on local regions (patches) rather than global image structure. By operating at the patch level, the model can better highlight spatially localized defects and achieve higher pixel-level AUROC. Extending MIL to the full image in a one-shot setting would not be meaningful, as global representations are insufficiently diverse to capture anomalous variations.  
\subsection{MVTecAD Results}
Table~\ref{tab:mvtec_results} summarizes the class-wise AUROC scores on the MVTecAD dataset, comparing the baseline PromptAD and WinCLIP against our proposed deviation-based method. Improvements achieved comparing to our baseline are highlighted in \textbf{bold}. 
\begin{figure}[h!]
    \centering
    \includegraphics[width=\linewidth]{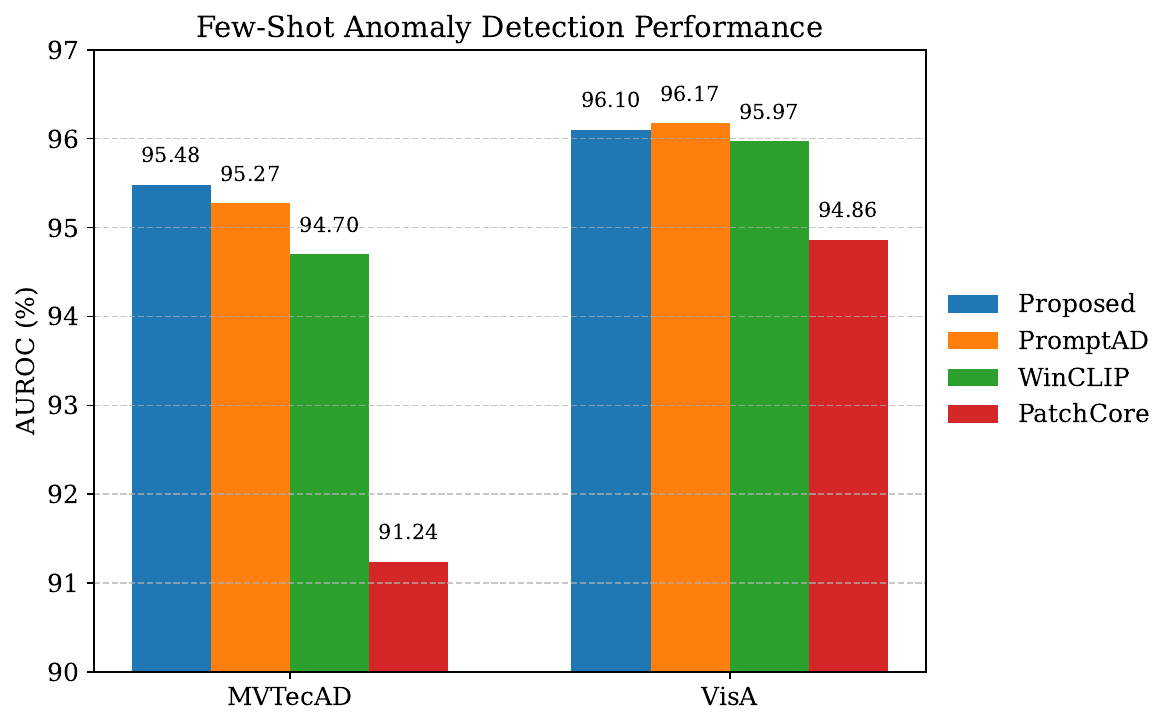}  
    \caption{Average AUROC comparison of PromptAD, WinCLIP, PatchCore, and the proposed method across MVTecAD and VISA datasets.}
    \label{fig:avg_comparison}
    \vspace{2pt}
\end{figure}
\paragraph{Analysis.}  
The results indicate that deviation-based patch scoring is particularly beneficial for categories with \emph{localized defects}. Classes such as 
\texttt{Capsule}, \texttt{Pill}, and \texttt{Metal Nut}, which contain cracks, contaminations, or missing regions, exhibit noticeable improvements due to the Top-K MIL mechanism emphasizing the most anomalous patches. Similarly, performance gains in \texttt{Zipper} and \texttt{Hazelnut} suggest that fine-grained texture irregularities are effectively distinguished from normal patterns.  

Minor reductions in AUROC for \texttt{Cable}, \texttt{Screw}, and \texttt{Wood} are observed. These categories contain anomalies that are either globally distributed or exhibit high intra-class variability, which are less sensitive to patch-level deviation modeling. Categories such as \texttt{Carpet}, \texttt{Grid}, and \texttt{Leather} maintain stable performance, indicating that both the baseline and our method have already saturated for simpler anomaly patterns.

\subsection{VISA Dataset Results}

Table~\ref{tab:visa_results} presents the segmentation results on the VISA dataset. Bold entries highlight improvements introduced by the deviation-based scoring.  

\paragraph{Analysis.}  
The proposed deviation-based scoring consistently improves the performance our model in categories with spatially concentrated anomalies, such as \texttt{Cashew}, \texttt{Chewinggum}, and \texttt{Pipe\_Fryum}. Notably, \texttt{PCB1} also benefits from patch-level deviation, demonstrating that structured defects, such as scratches or missing circuitry, can be accurately localized. 
\begin{table}[h!]
\centering
\caption{Impact of the deviation coefficient $\lambda$ (VISA, AUROC \%).}
\label{tab:alpha_ablation}
\begin{tabular}{lccc}
\hline
Class & Proposed & $\lambda=0.1$ & $\lambda=0.01$ \\
\hline
Candle & 94.69 & 94.43 & 94.57 \\
Capsules & 94.01 & 94.00 & 94.00 \\
Cashew & 98.88 & 98.83 & 98.86 \\
Chewinggum & 99.25 & 99.23 & 99.25 \\
Fryum & 95.63 & 95.77 & 95.49 \\
Macaroni1 & 97.73 & 97.72 & 97.76 \\
Macaroni2 & 95.49 & 95.32 & 95.31 \\
PCB1 & 95.83 & 95.72 & 95.73 \\
PCB2 & 93.34 & 93.93 & 94.58 \\
PCB3 & 94.65 & 94.53 & 94.59 \\
PCB4 & 94.49 & 94.46 & 94.31 \\
Pipe\_Fryum & 99.19 & 99.14 & 99.17 \\
\hline
\end{tabular}
\vspace{3pt}
\end{table}
Slight reductions in AUROC for \texttt{PCB2} and \texttt{PCB4} suggest that anomalies dispersed across fine circuit patterns are less sensitive to patch-level deviations and may require global structural reasoning. Categories with simpler or more uniform defect patterns, such as \texttt{Capsules}, \texttt{PCB3}, and \texttt{Macaroni1}, maintain robust performance, confirming the method’s generalization across defect types.
\paragraph{Overall Performance Comparison.}  
To provide a comprehensive view of the method’s effectiveness, Figure~\ref{fig:avg_comparison} presents the average AUROC of PromptAD, WinCLIP, PatchCore, and the proposed approach across both MVTecAD and VISA datasets. As illustrated, the proposed method achieves the best performance on MVTecAD, indicating its strong ability to capture fine-grained, localized anomalies. On the VISA dataset, it maintains competitive results, closely matching or exceeding the baseline methods. This comparative visualization underscores that integrating deviation-guided scoring with prompt-based representations leads to consistent performance gains across benchmarks with diverse anomaly patterns.

\hl{It is seen from Tables }\ref{tab:mvtec_results}\hl{, and }\ref{tab:visa_results}\hl{, that the improvements over PromptAD are numerically small, but the consistent behavior across datasets and metrics in the few-shot setting reflects improved robustness and generalization without added supervision or model complexity.}

\subsection{Sensitivity Analysis}  
We further conduct a sensitivity analysis to examine how variations in key hyperparameters affect the performance of our model. In particular, we analyze the impact of the deviation coefficient $\lambda$, the Top-K percentage parameter and the impact of the confidence parameter $a$ used in the MIL-based deviation scoring. This analysis provides insight into the robustness of the proposed method and helps identify stable operating ranges for different anomaly distributions.
\begin{table}[h!]
\centering
\caption{Impact of Top-K$\%$ (VISA, AUROC \%).}
\label{tab:k_ablation}
\begin{tabular}{lccc}
\hline
Class & Proposed & $K=20\%$ & $K=30\%$ \\
\hline
Candle & 94.69 & 94.59 & 94.75 \\
Capsules & 94.01 & 94.00 & 94.01 \\
Cashew & 98.88 & 98.79 & 98.93 \\
Chewinggum & 99.25 & 99.26 & 99.24 \\
Fryum & 95.63 & 95.54 & 95.49 \\
Macaroni1 & 97.73 & 97.77 & 97.73 \\
Macaroni2 & 95.49 & 95.27 & 95.14 \\
PCB1 & 95.83 & 95.72 & 95.17 \\
PCB2 & 93.34 & 94.43 & 94.45 \\
PCB3 & 94.65 & 94.64 & 94.58 \\
PCB4 & 94.49 & 94.11 & 94.06 \\
Pipe\_Fryum & 99.19 & 99.18 & 99.1 \\
\hline
\end{tabular}
\vspace{5pt}
\end{table}
\begin{table}[h!]
\centering
\caption{Impact of confidence parameter $a$ (VISA, AUROC \%).}
\resizebox{\linewidth}{!}{
\label{tab:alpha_ablation}
\begin{tabular}{lccccc}
\hline
Class & $a=1$ & $a=3$ & $a=5$ & $\a=7$ & $a=9$ \\
\hline
Candle & 94.38 & 94.49 & 94.69 & 94.7 & 94.65 \\
Capsules & 93.47&93.98 &94.01 & 94 & 94 \\
Cashew & 98.43&98.72 &98.88 & 98.87 & 98.84 \\
Chewinggum &99.1 &99.16&99.25 & 99.26 & 99.22 \\
Fryum & 94.38& 95.01&95.63 & 95.61 & 95.59 \\
Macaroni1 & 97.08&97.58&97.73 & 97.71 & 97.68 \\
Macaroni2 & 94.7&95.1&95.49 & 95.46 & 95.39 \\
PCB1 & 95.3&95.71&95.83 & 95.79 & 95.72 \\
PCB2 & 92.82&93.11&93.34 & 93.35 & 93.27 \\
PCB3 & 94.68&94.62&94.65 & 94.64 & 94.58 \\
PCB4 & 94.31&94.39&94.49 & 94.45 & 94.37 \\
Pipe\_Fryum &99.01 &99.12&99.19 & 99.17 & 99.14 \\
\hline
\end{tabular}
}
\vspace{5pt}
\end{table}
\paragraph{Impact of Deviation Coefficient $\lambda$.}  To evaluate the influence of the deviation coefficient $\lambda$ on the performance of the proposed method,, we conduct analayze on the VISA dataset. Table~\ref{tab:alpha_ablation} reports class-wise AUROC scores for three different values of $\lambda$.  

The results indicate that the method is relatively stable across different $\lambda$ values. For categories with more globally distributed anomalies, such as \texttt{PCB2}, a smaller deviation coefficient ($\lambda=0.01$) slightly improves segmentation by reducing over-penalization of moderate deviations.
\paragraph{Impact of Top-K Parameter.}  We further investigate the effect of the Top-K parameter on the performance of the proposed method, as reported in Table~\ref{tab:k_ablation}. A higher $K$ increases the number of patch scores considered when computing the MIL-based deviation score. Specifically, $K$ is defined as a percentage of the total number of patches in an image (e.g., $K = 10\%$ or $20\%$), meaning that only the top $K$ highest deviation scores are used for final anomaly scoring.

Increasing $K$ generally improves performance for categories where anomalies are distributed across multiple regions, as observed in \texttt{PCB2}. Conversely, categories with highly localized defects, such as \texttt{Fryum} or \texttt{Pipe\_Fryum}, exhibit limited sensitivity to $K$, demonstrating that the method is robust to this hyperparameter in such scenarios. \hl{Importantly, this behavior does not assume prior knowledge of anomaly characteristics, rather, it reflects intrinsic defect patterns commonly encountered in industrial inspection, where the spatial extent of anomalies is governed by the underlying manufacturing process.}
\paragraph{Impact of the confidence parameter $a$.} The parameter $a$ defines the minimum deviation margin between normal and anomalous samples during training. We evaluate $a \in \{1, 3, 5, 7, 9\}$ on the VISA dataset. It is seen from Table \ref{tab:alpha_ablation} that small values of $a$ lead to insufficient separation, while performance stabilizes once a meaningful margin is reached, with $a = 5$ providing the best trade-off between accuracy and stability.
\section{Conclusion}
\label{conclusion}
Few-shot anomaly detection remains challenging due to the need for accurate localization under extremely limited supervision. In this work, we proposed a \textbf{deviation-guided prompt learning framework} that integrates vision--language alignment with statistically grounded patch-level anomaly scoring. By introducing \textbf{learnable context vectors} into textual prompts and optimizing a \textbf{Top-K MIL-driven deviation loss}, the proposed method enhances discrimination between normal and anomalous regions while preserving interpretability. Experiments on \textbf{MVTecAD} and \textbf{VISA} demonstrate that the proposed approach consistently improves pixel-level anomaly detection, particularly for localized and texture-rich defects, while remaining stable across a wide range of hyperparameter settings. These results indicate that combining prompt-based global semantics with deviation-based local scoring is well suited for industrial anomaly detection scenarios.

Future work may extend this framework to \textbf{video-based anomaly detection}, explore \textbf{overlapping patch designs} within CLIP to improve spatial sensitivity, or incorporate more structured priors into prompt construction.



\section*{Acknowledgments}
Work performed while M. Poudineh was an intern at CRIM. The internship was financed by CRIM with support from the Ministry of Economy, Innovation, and Energy (MEIE) of the Government of Quebec.

\bibliographystyle{apalike}  
\bibliography{example} 

\end{document}